\documentclass[conference]{IEEEtran}
\IEEEoverridecommandlockouts

\usepackage{cite}
\usepackage{amsmath,amssymb,amsfonts}
\usepackage{algorithmic}
\usepackage{graphicx}
\usepackage{textcomp}
\usepackage{xcolor}
\usepackage{url}
\usepackage{breakurl}
\usepackage{multirow}

\def\BibTeX{{\rm B\kern-.05em{\sc i\kern-.025em b}\kern-.08em
    T\kern-.1667em\lower.7ex\hbox{E}\kern-.125emX}}
\begin{document}

\title{V-CAS: A Realtime Vehicle Anti Collision System Using Vision Transformer on Multi-Camera Streams}

\author{\IEEEauthorblockN{Muhammad Waqas Ashraf\textsuperscript{ c}}
\IEEEauthorblockA{\textit{College of EME, NUST}\\
Pakistan \\
washraf.ce22ceme@student.nust.edu.pk}
\and
\IEEEauthorblockN{Ali Hassan}
\IEEEauthorblockA{\textit{College of EME, NUST}\\
Pakistan \\
alihassan@ceme.nust.edu.pk}
\and
\IEEEauthorblockN{Imad Ali Shah}
\IEEEauthorblockA{\textit{University of Galway}\\
 Ireland \\
i.shah2@universityofgalway.ie}
}

\maketitle

\begin{abstract}
This paper introduces a robust real-time Vehicle Collision Avoidance System (V-CAS) aimed at enhancing vehicle safety through environmental perception-based adaptive braking. V-CAS utilizes the advanced vision-based transformer model RT-DETR, DeepSORT tracking, speed estimation, brake light detection, and an adaptive braking mechanism. It computes a composite collision risk score from vehicles' relative accelerations, distances, and detected braking actions, leveraging brake light signals and trajectory data through multiple camera streams for improved scene perception. Implemented on the Jetson Orin Nano, V-CAS enables real-time collision risk assessment and proactive mitigation via adaptive braking. A comprehensive training process was conducted on various datasets for comparative analysis, followed by fine-tuning the selected object detection model using transfer learning. The system's effectiveness was rigorously evaluated on the Car Crash Dataset (CCD) from YouTube and through real-time experiments, achieving over 98\% accuracy with an average proactive alert time of 1.13 seconds. Results show significant improvements in object detection and tracking, enhancing collision avoidance compared to traditional single-camera methods. This research highlights the potential of low cost, multi-camera embedded vision transformer systems to advance automotive safety through enhanced environmental perception and proactive collision avoidance mechanisms.
\end{abstract}


\section{Introduction}

The increase in car ownership, driven by economic growth and the desire for convenience, has resulted in a rise in traffic accidents, leading to significant loss of life. Research shows that approximately 77\% of these accidents are caused by drivers \cite{b1}. This concerning trend underscores the urgent need for intelligent road safety systems that can perceive surrounding traffic objects and prevent collisions. These systems utilize various data sources, including vehicle speed, accelerometers, and video feeds. Recent advancements have seen researchers incorporating Light Detection and Ranging (LiDAR) sensor inputs and monocular camera images to enhance the performance of collision avoidance systems. 

ADAS can be categorized into two main types: (1) Passive Safety focuses on reducing injuries during a crash through high production safety standards, while (2) Active Safety systems proactively prevent accidents by using sensors such as radar, cameras, and ultrasonic devices to detect potential hazards like nearby vehicles or sudden braking. When a threat is identified, these systems alert the driver with visual or audio warnings or initiate automatic braking to avert collisions.  Modern systems often integrate cameras and radars, providing distinct advantages. However, the addition of sensors can increase vehicle costs and design complexity. To address this, researchers are investigating computer vision insights, particularly in object detection (OD) techniques that utilize either depth-based or camera-based sensors. 

The proposed system utilizes spatial feature extraction from RGB feeds captured by three cameras, facilitating enhanced scene interpretation and a broader field of view (FOV). It integrates object detection and tracking algorithms to predict collision scores based on relative motion, all executed efficiently in real-time on edge devices like the Jetson Orin Nano. This method promises a more robust and computationally efficient recognition of surrounding traffic objects. The structure of this paper is as follows: Section II reviews related work in the field, Section III details the methodology of the proposed model, and Sections IV and V present the experimentation results and conclusions, along with future directions for research.

\section{Related Work}

Existing approaches for collision prediction and avoidance systems can be broadly categorized into three main groups: (1) motion trajectory prediction-based models using Deep Reinforcement Learning (DRL), (2) radar-camera sensor fusion techniques, and (3) vision-based approaches with Deep Learning (DL).

\subsection{Motion Trajectory Prediction using Deep Reinforcement Learning}
An efficient collision detection system relies on confident prediction of vehicle motion and trajectory. Lefevre et al. \cite{b2} explores various motion prediction approaches, and categorized motion prediction models into three main types: physics-based, manoeuvre-based and interaction-aware. DRL, where algorithms learn from trial and error, has shown promise in navigation systems. Kahn et al. \cite{b3} propose a collision avoidance mechanism utilizing a standard stereo camera, where their navigation model outperformed Double Q-learning in achieving fully autonomous navigation. Chen et al. \cite{b4} shows a decentralized collision avoidance algorithm using DRL that predicts best paths with minimal collision risk, considering the positions and velocities of surrounding vehicles.  However, limitations do exist as these algorithms require vast amounts of training data and limit real-world generalizability. Additionally, their computational demands can lead to delays in critical moments, making them hard to be implemented in real-time for passenger vehicles.

\subsection{Radar-Camera Sensor Fusion Approaches}
Several DL architectures have been proposed for radar-camera and LiDAR sensor fusion in collision avoidance systems. Radar offers all-weather functionality, however, detailed information about object size and shape is absent. Camera-based sensors provide rich visual data like lane markings, traffic signals, and object shapes. LiDAR (Light Detection and Ranging) creates a 3D point cloud representation of the environment, offering precise distance and shape information but is costly.  Some of the fusion approaches are: Early Fusion which merges raw radar data and camera images at the beginning of the network and processed by a single DL model as proposed by Xu et al. \cite{b5}. This approach is computationally efficient but requires careful pre-processing. Late Fusion separates DL models process for radar and camera data independently, extracting features which are then fused at a later stage for final decision-making. Kim et al. \cite{b6} proposed late fusion of camera with LiDAR data for pedestrian detection. This approach allows for independent optimization of each sensor model but might lose some information. Feature Level Fusion involves processing sensors through individual feature extraction layers which are then concatenated before feeding them into a final classification layer. Zhu et al. \cite{b7} introduced feature level fusion. Attention-based Fusion focuses on the most relevant features, dynamically allocating weights for improved robustness but is computationally expensive as seen in the proposed method by Huang et al. \cite{b8} for vehicle detection.

\subsection{Vision-based Approaches with Deep Learning (DL)}
While DRL and sensor fusion techniques show their robustness, vision-based approaches offer a promising alternative due to their cost-effectiveness and ease of integration.  Monocular vision proved to be valuable by estimating Time-to-Collide (TTC) and addresses the issue of collision rarity. Methods for TTC estimation, including those by Shi et al. \cite{b9}, involve feature tracking, motion divergence analysis, and optical flow. These approaches have limitations like lack of hardware efficiency, robustness, and reliance on additional data like lane markings. The possibility of enhancing object detection accuracy across multiple cameras by gathering a more holistic view of the scene was highlighted as future work by Sharma \cite{b10}. Datondji et al. \cite{b11} has proposed YOLO, a DL model, for vehicle recognition and tracking in traffic videos. Vehicle detection achieved higher accuracy in classifying and counting vehicles across various highway videos \cite{b12}. Researchers, like Ngeni et al. \cite{b13},  have used YOLO variants with DeepSORT for real-time traffic tracking related models. Similarly, various models of SSD models like MobileNets v1 to v3 \cite{b14} were also used for real time object detection specially for low power embedded devices, however, their detection performance in terms of  accuracy was not very promising. Transformers, on the other hand, showed promising results in terms of contextual relationship and accuracy for NLP but, due to their lack of real-time ability, were not being used in vision related tasks.

RT-DETR \cite{b15}, proposed by Zhao et al in CVPR 2024,  proved their vision-based transformer model as the new state-of-the-art (SOTA) real-time OD model where it has beaten YOLOs in performance and speed. Many leading autonomous car manufacturers like Tesla and Kia have also shifted towards totally vision based object detection systems \cite{b16}. Tesla's Full Self-Driving (FSD) software relies solely on cameras, abjuring radar and lidar \cite{b17}. Keeping in view the cost effectiveness, ease of integration, and optimal performance for real-time predictions, our focus is towards vision-based methods acquiring input data from multiple camera streams. 

\begin{figure}[htbp]
\centerline{\includegraphics[width=0.48\textwidth]{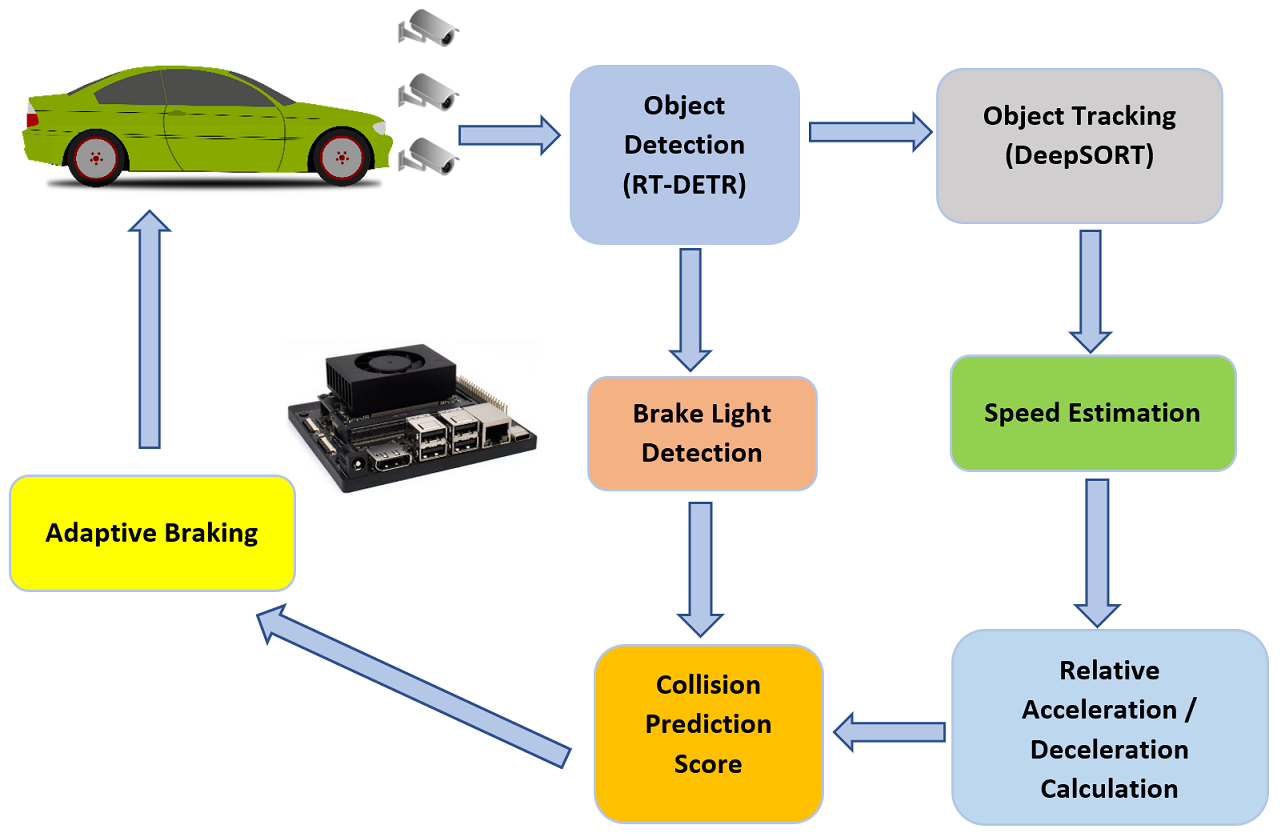}}
\caption{Basic Architecture of VCAS}
\label{fig}
\end{figure}

\section{PROPOSED METHODOLOGY}
This section presents an overview of the proposed system, with key building blocks and their integration. Fig 1 shows the block diagram of our proposed system V-CAS architecture. An array of 3 cameras was used, forming a wider FOV. A real-time vision-based transformer model RT-DETR was used to detect moving or static objects like vehicles, pedestrians etc. DeepSORT is a tracking method which combines the predictive power of the Kalman filter with the robustness of DL. Comparing the positions of each tracked object, their speeds and ultimately rate of acceleration were estimated. Resultantly, a collision score is assigned for each object. All these calculations were carried out on Jetson Orin Nano. Once the collision score crosses a specific threshold, the Jetson device sends a braking signal via its 40-pin expansion header to the vehicle's adaptive braking mechanism, applying braking.

\subsection{Objects Detection – RT-DETR}
The V-CAS backbone utilizes a vision-based transformer model, RT-DETR, that competes with single-stage, real-time OD models such as YOLO latest variants. This model is pretrained on the COCO dataset and fine-tuned on several custom datasets, incorporating only the necessary parameters and layers for classifying the desired classes through transfer learning. This approach strikes a balance between speed and accuracy, making it well-suited for our deployment needs.

\subsection{Objects Tracking – DeepSORT}
One of the renowned multi-object tracker, DeepSORT, was used. It combines the strengths of DL for feature extraction and a classic Kalman filter for data association. It relies on two primary modules: (1) Deep Appearance Descriptor using a pre-trained deep convolutional neural network (CNN) to extract high-level features in each video frame and (2) Kalman Filter and Hungarian Algorithm. Kalman filter is used to predict the state of each detected object across frames and helps in keeping track even during occlusions or missed detections. The Hungarian algorithm is employed to associate detections in the current frame with existing tracks or initiate new ones based on similarity between predicted states and current detections.

\subsection{Speed Estimation}
To estimate the speed of detected objects, the system calculates the Euclidean distance between the object's positions in consecutive frames using the distance formula  (Equation 1). Traveling distance of the vehicle between these two frames called pixel displacement $(\Delta d)$ is represented by: 

\begin{equation}
\Delta d = \sqrt{((x_{i+1} - x_i)^2 + (y_{i+1} - y_i)^2)} \qquad   
\end{equation}

where $(x_i, y_i)$ and $(x_{i+1}, y_{i+1})$ are the horizontal and vertical pixel position of the target vehicle on frame $i$ and $i+1$ respectively. Pixel displacement is then converted to real-world meters using pre-calibrated pixel-per-meter (ppm), which was 20 in our case. Finally, the function calculates the speed estimation (Equation 2)  as:

\begin{equation}
  v = \frac{\Delta d}{ppm} \times time_{const} \times 3.6
\end{equation}

\subsection{Calculating Relative Rate of Acceleration}
The relative rate of acceleration (Equation 3) for each tracked object which crosses the invisible grid around the subject vehicle is calculated to assess collision risk. A queue of 20 speed values is maintained for each object, divided into initial and final buffers of 10 values each. The relative acceleration is then divided by a variable $\beta$ which was $0.0625$ in our case, shown mathematically as:

\begin{equation}
a = \frac{\frac{1}{10} \sum_{i=1}^{10} final\_speed_i - \frac{1}{10} \sum_{j=1}^{10} initial\_speed_j}{20 \times \beta}
\end{equation}

\subsection{Brake Light Detection}
Although speed estimation provides a basic method for predicting collisions, it comes with certain drawbacks. (1) The whole collision prediction calculation on a single pixel’s relative displacement is not very robust in some cases. (2) The failure in detecting our objects of interest present on the road, more likely at night time, will result in no collision prediction at all. Therefore, another widely used method was integrated into V-CAS, i.e to predict the collision by detecting the forward vehicle brake light to caution the system that front moving vehicle is coming to a halt and may collide with the host vehicle within trajectory of movement. In this way, at night time or in bad weather condition, even if the vehicles remain undetected by OD model, the brake light is detected easily giving an additional check that there is a vehicle in front and that is halting too. Moreover, it triggers an alert when the detected brake light of a forward-moving car is close to the host vehicle, skipping the speed estimation and engaging emergency brakes at once as needed.

\begin{table*}[htbp]
\centering
\caption{System Resources for Training and Inference}
\begin{tabular}{|l|l|p{4.7cm}|p{8cm}|}
\hline
\textbf{Resource} & \textbf{Object} & \textbf{Training Specification} & \textbf{Inference Specifications} \\
\hline
Hardware & CPU & Intel i9-13900K & 6-Core ARM Cortex-A78AE v8.2 64-Bit \\
 & GPU & Nvidia Geforce RTX 4090Ti 24GB & 1024-Core Nvidia Ampere Architecture GPU with 32 Tensor Cores \\
 & RAM & 64GB & 8GB 128-Bit LPDDR5 \\
 & Power & 450W & 15W \\
\hline
Software & OS & Windows 11 64-Bit & JetPack 6.0 Developer Ubuntu 22.04 \\
 & Framework & \begin{tabular}[t]{@{}l@{}}Pytorch 2.3.1, CUDA 12.1, cuDNN 9.2.1\end{tabular} & Pytorch 2.2.0, CUDA 12.2.12, cuDNN 8.9.4, TensorRT 8.6.2 \\
\hline
\end{tabular}
\end{table*}

\begin{table*}[htbp]
\centering
\caption{Details of Training Hyper-Parameters and Data Sets Used for OD Model : RT-DETR}
\begin{tabular}{|l|c|c|c|c|c|c|c|c|c|c|}
\hline
\textbf{Dataset} & \textbf{\begin{tabular}[c]{@{}c@{}}Training\\Set\end{tabular}} & \textbf{\begin{tabular}[c]{@{}c@{}}Validation\\Set\end{tabular}} & \textbf{\begin{tabular}[c]{@{}c@{}}Classes\end{tabular}} & \textbf{\begin{tabular}[c]{@{}c@{}}Epochs\end{tabular}} & \textbf{\begin{tabular}[c]{@{}c@{}}Batch\\ Size\end{tabular}} & \textbf{\begin{tabular}[c]{@{}c@{}}Image\\ Size\end{tabular}} & \textbf{Optimizer} & \textbf{\begin{tabular}[c]{@{}c@{}}Learning\\ Rate\end{tabular}} & \textbf{Momentum} & \textbf{Decay} \\
\hline
Vehicle i2 & 6,638 & 820 & 25 & 100 & 32/16 & 640 & AdamW & 0.0003 & 0.9 & 0.0005 \\
% \hline
\begin{tabular}[c]{@{}l@{}}Brake Light Detection\end{tabular} & 18,939 & 3,586 & 2 & 100 & 32/16 & 640 & SGD & 0.01 & 0.5 & 0.005 \\
\hline
\end{tabular}
\end{table*}

\subsection{Fusion of Multiple Camera Sensors}
There are few known video pipelines and SDK like Nvidia DeepStream \cite{b18} and GStreamer \cite{b19} for real-time multiple camera streams fusion, however, they are very difficult to integrate and not flexible enough to be modified for custom application easily. Therefore, a more simplistic and flexible approach was adopted using OpenCV, Numpy and multi-threading. The ingeniously created ‘Vstream’ class is designed to handle individual video streams, continuously capturing frames from different sources in separate threads for efficient processing. These frames are resized to a uniform dimension and stored in a thread-safe manner. Then all camera frames are read simultaneously and they are concatenated into a single frame. This combined frame is then passed to our OD model. The detections are subsequently passed on to single tracker to maintain consistent identities of objects across frames. Finally, bounding boxes and labels are drawn on the combined window to indicate tracked objects. This approach enables real-time fusion and processing of multiple video streams in a simple way.

\subsection{Calculating Collision Prediction Score}
The relative rate of acceleration of the same object against its earlier values from host vehicle plays a pivotal role in our collision avoidance system. If it is increasing in the same trajectory, it means that our vehicle and the detected object are closing in and vice versa. Depending upon these values, a confidence score was assigned. If it crosses a threshold $(>60\%)$, then an electric signal is generated from 40 pin Expansion Header of Jetson device to the braking mechanism as pulse width modulated (pwm) signal. Where width of the pulse is proportional to the confidence score of collision prediction. It displays collision warnings on the monitor screen to the driver. Additionally, the custom trained Brake Light detection model also keeps on detecting the vehicles in the scene with brakes “ON” status. If any such vehicle comes into near proximity of the host vehicle in the same or cross-sectional trajectory, it generates an emergency braking signal from our embedded device to the vehicle braking system.

% \begin{table*}[htbp]
% \centering
% \caption{Performance Evaluation of Latest Real-Time Object Detectors on The Vehicles i2 Public Dataset}
% \begin{tabular}{|l|c|c|c|c|c|c|c|}
% \hline
% \multirow{\textbf{Model}} & \multirow{\textbf{Size (MB)}} & \multirow{\textbf{Parameters (Mn)}} & \multirow{\begin{tabular}[c]{@{}c@{}}\textbf{Inference Time}\\ \textbf{per Image (ms)}\end{tabular}} & \multicolumn{4}{c|}{\textbf{Evaluation Metrics}} \\
% \cline{5-8}
%  &  &  &  & \textbf{Precision} & \textbf{Recall} & \textbf{mAP50} & \textbf{mAP50-95} \\
% \hline
% YOLOv8s & 21.4 & 11.13 & \textbf{0.7} & 0.844 & 0.747 & 0.830 & 0.706 \\
% YOLOv8m & 49.6 & 25.85 & 1.8 & 0.767 & 0.770 & 0.816 & 0.697 \\
% YOLOv8l & 83.5 & 43.62 & 2.7 & 0.771 & 0.785 & 0.803 & 0.688 \\
% YOLOv9s & \textbf{14.5} & \textbf{7.29} & 0.9 & \textbf{0.864} & 0.731 & 0.842 & 0.706 \\
% YOLOv9c & 49.2 & 25.34 & 2.6 & 0.705 & 0.838 & 0.842 & 0.714 \\
% YOLOv9e & 111.0 & 57.39 & 5.9 & 0.828 & 0.723 & 0.782 & 0.664 \\
% YOLOv10s & 15.7 & 8.05 & 1.0 & 0.850 & 0.736 & \textbf{0.861} & \textbf{0.730} \\
% YOLOv10m & 31.9 & 16.47 & 1.9 & 0.856 & 0.706 & 0.801 & 0.676 \\
% YOLOv10b & 39.5 & 20.45 & 2.5 & 0.750 & 0.721 & 0.790 & 0.654 \\
% RT-DETR L & 63.1 & 32.03 & 2.6 & 0.857 & \textbf{0.845} & 0.853 & 0.724 \\
% \hline
% \end{tabular}
% \end{table*}

\begin{table*}[htbp]
\centering
\caption{Performance Evaluation of Latest Real-Time Object Detectors on The Vehicles i2 Public Dataset}
\begin{tabular}{|l|c|c|c|c|c|c|c|}
\hline
\textbf{Model} & \textbf{Size (MB)} & \textbf{Parameters (Mn)} & \textbf{Inference Time per Image (ms)} & \textbf{Precision} & \textbf{Recall} & \textbf{mAP50} & \textbf{mAP50-95} \\
\hline
YOLOv8s & 21.4 & 11.13 & \textbf{0.7} & 0.844 & 0.747 & 0.830 & 0.706 \\
YOLOv8m & 49.6 & 25.85 & 1.8 & 0.767 & 0.770 & 0.816 & 0.697 \\
YOLOv8l & 83.5 & 43.62 & 2.7 & 0.771 & 0.785 & 0.803 & 0.688 \\
YOLOv9s & \textbf{14.5} & \textbf{7.29} & 0.9 & \textbf{0.864} & 0.731 & 0.842 & 0.706 \\
YOLOv9c & 49.2 & 25.34 & 2.6 & 0.705 & 0.838 & 0.842 & 0.714 \\
YOLOv9e & 111.0 & 57.39 & 5.9 & 0.828 & 0.723 & 0.782 & 0.664 \\
YOLOv10s & 15.7 & 8.05 & 1.0 & 0.850 & 0.736 & \textbf{0.861} & \textbf{0.730} \\
YOLOv10m & 31.9 & 16.47 & 1.9 & 0.856 & 0.706 & 0.801 & 0.676 \\
YOLOv10b & 39.5 & 20.45 & 2.5 & 0.750 & 0.721 & 0.790 & 0.654 \\
RT-DETR L & 63.1 & 32.03 & 2.6 & 0.857 & \textbf{0.845} & 0.853 & 0.724 \\
\hline
\end{tabular}
\end{table*}

\begin{table*}[htbp]
\centering
\caption{Comparison of Different Real-Time Object Detectors on the Brake Light Detection Dataset}
\label{tab:model_comparison}
\resizebox{\textwidth}{!}{%
\begin{tabular}{|l|c|c|c|c|c|c|c|c|c|c|c|c|c|c|}
\hline
\textbf{Model} & \textbf{Size (MB)} & \textbf{GFLOPS} & \multicolumn{6}{c|}{\textbf{Brake OFF Class (-ve class)}} & \multicolumn{6}{c|}{\textbf{Brake ON Class (+ve class)}} \\
\cline{4-15}
 &  &  & \textbf{TP} & \textbf{FP} & \textbf{FN} & \textbf{Precision} & \textbf{Recall} & \textbf{F1-Score} & \textbf{TP} & \textbf{FP} & \textbf{FN} & \textbf{Precision} & \textbf{Recall} & \textbf{F1-Score} \\
\hline
YOLOv8s & 21.4 & 28.5 & 1479 & 1308 & 640 & 0.530 & 0.697 & 0.604 & 1278 & 1026 & 586 & 0.554 & 0.685 & 0.614 \\
YOLOv8m & 49.6 & 78.8 & 1560 & 1238 & 559 & 0.557 & 0.736 & 0.634 & 1350 & 1007 & 514 & 0.572 & 0.724 & 0.640 \\
YOLOv8l & 83.5 & 164.9 & 1572 & 1245 & 547 & 0.558 & 0.741 & 0.637 & 1328 & 996 & 536 & 0.571 & 0.712 & 0.635 \\
YOLOv9s & \textbf{14.5} & 26.7 & 1575 & 1282 & 544 & 0.551 & 0.743 & 0.633 & 1325 & 1028 & 539 & 0.563 & 0.710 & 0.629 \\
YOLOv9c & 49.2 & 102.4 & 1597 & 1240 & 522 & 0.562 & 0.753 & 0.646 & 1337 & 962 & 527 & 0.581 & 0.717 & 0.643 \\
YOLOv9e & 111.0 & 189.2 & 1596 & 1249 & 523 & 0.560 & 0.753 & 0.646 & 1361 & 961 & 503 & 0.586 & 0.730 & 0.650 \\
YOLOv10s & 15.7 & \textbf{24.5} & 1528 & 1163 & 591 & 0.567 & 0.721 & 0.636 & 1307 & 867 & 557 & 0.601 & 0.701 & 0.647 \\
YOLOv10m & 31.9 & 63.6 & 1474 & 1321 & 645 & 0.527 & 0.696 & 0.599 & 1366 & 989 & 498 & 0.580 & 0.732 & 0.648 \\
YOLOv10b & 39.5 & 98.1 & 1391 & 1357 & 728 & 0.506 & 0.656 & 0.573 & 1362 & 1020 & 502 & 0.572 & 0.730 & 0.642 \\
RT-DETR L & 63.1 & 103.4 & \textbf{1629} & \textbf{850} & \textbf{490} & \textbf{0.657} & \textbf{0.769} & \textbf{0.710} & \textbf{1373} & \textbf{631} & \textbf{491} & \textbf{0.685} & \textbf{0.736} & \textbf{0.710} \\
\hline
\end{tabular}%
}
\end{table*}

\section{EXPERIMENTATION}
\subsection{Datasets for Object Detection}
Two publicly available datasets from Roboflow were used to fine-tune the pre-trained RT-DETR-L object detection model: (1) Vehicle i2 dataset having a total of 7458 images of size 640 x 640. As preprocessing step, auto-orientation of pixel data was applied with EXIF-orientation stripping. It has 25 different classes having covered almost all categories of vehicles from ambulance to bus, rickshaw, bikes and another potential collision object i.e pedestrians as well. (2) Brake Light Detection Dataset having a total of 22,525 images of size 640 x 640. As preprocessing step, auto-orientation of pixel data was applied with EXIF-orientation stripping. Additional augmentations were applied as well: (a) 50\% horizontal flip, (b) Random crop between 0 and 20 percent of the image, (c) Random brightness between -25 and +25 percent, (d) Random Gaussian blur between 0 and 1.5 pixels and (e) Salt and pepper noise to 5 percent of pixels. It is a binary class dataset having classes ‘Brake Off’ and ‘Brake ON’.

\subsection{Datasets for Collision Avoidance Evaluation}
To the best of our knowledge, there is currently no publicly available video dataset of paired multi-camera streams for traffic data and collision prediction analysis. Therefore, we used hybrid datasets: (1) Our own, recorded from three vehicle mounted cameras for more than 10 hours recorded over the highways and city areas under normal and rash driving conditions in day, clouds, dusk and night with the car speed between 10 to 120 km/hr. (2) Car Crash Dataset (CCD) \cite{b20} which is collected and published for traffic accident analysis. It contains real traffic accident videos collected from YouTube channels and split them to get 1,500 trimmed videos. Each video contains 50 frames with 10 frames per second. The 3,000 normal videos are randomly sampled from BDD100K dataset \cite{b21}.

\subsection{System Resources and Training Setup}
Hardware and software resources utilized for training and inference  are shown in Table I. Table II summarizes the training hyper-parameters and details for both datasets being used for our model fine tuning for the specific tasks. 

\subsection{Evaluation Metrics}
The spatial evaluation of our object detection system is primarily based on mean average precision (mAP). mAP50 refers to the model's accuracy at an Intersection over Union (IoU) threshold of 0.5, which reflects the overlap between the predicted and actual object bounding boxes. This provides a single metric indicating the model’s ability to correctly identify objects with a reasonable degree of overlap. On the other hand, mAP 50-95 is a more stringent evaluation, calculating the average precision over multiple IoU thresholds (from 0.5 to 0.95, in steps of 0.05), giving a comprehensive indication of the model's robustness and accuracy across varying degrees of overlap. Additionally, to assess the object detection and collision prediction performance on the Vehicle i2, Brake Light Detection, and CCD datasets, a confusion matrix is used to compute key metrics like Recall (Degree of Completeness) and Precision (Degree of Correctness). This matrix consists of parameters like true positives (TP), true negatives (TN), false positives (FP), and false negatives (FN), which categorize how well the model identifies and misidentifies objects. Among these, Recall is particularly crucial in our analysis, as it indicates the system's effectiveness in detecting potential objects and predicting collisions.

% \begin{equation}
% Accuracy = \frac{TP + TN}{TP + TN + FP + FN}
% \end{equation}

% \begin{equation}
% Precision = \frac{TP}{TP + FP}
% \end{equation}

% \begin{equation}
% Recall = \frac{TP}{TP + FN}
% \end{equation}

\subsection{Experimental Analysis}
Table III and IV shows a comparison of latest real-time OD models on our training datasets for vehicles and brake light detection respectively. We have selected only latest real-time OD models which have been launched after 2022. Those OD models which are good in accuracy but lacks in real-time capability like vision-based end to end object detection transformer (DETR) or Dino-DETR were not included.

\begin{figure}[htbp]
\centerline{\includegraphics[width=0.49\textwidth]{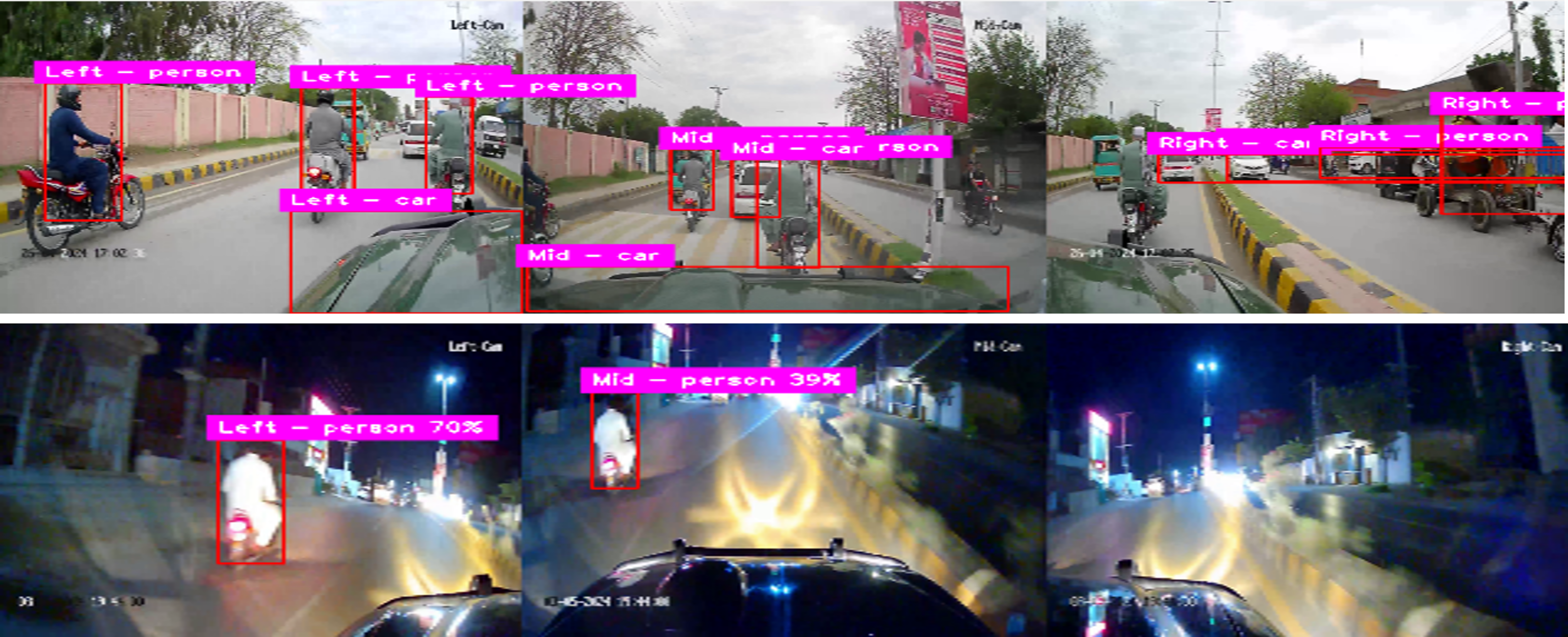}}
\caption{Multi-Camera Object Detection using V-CAS OD Model in Day and Night}
\label{fig}
\end{figure}

\begin{figure}[htbp]
\centerline{\includegraphics[width=0.49\textwidth]{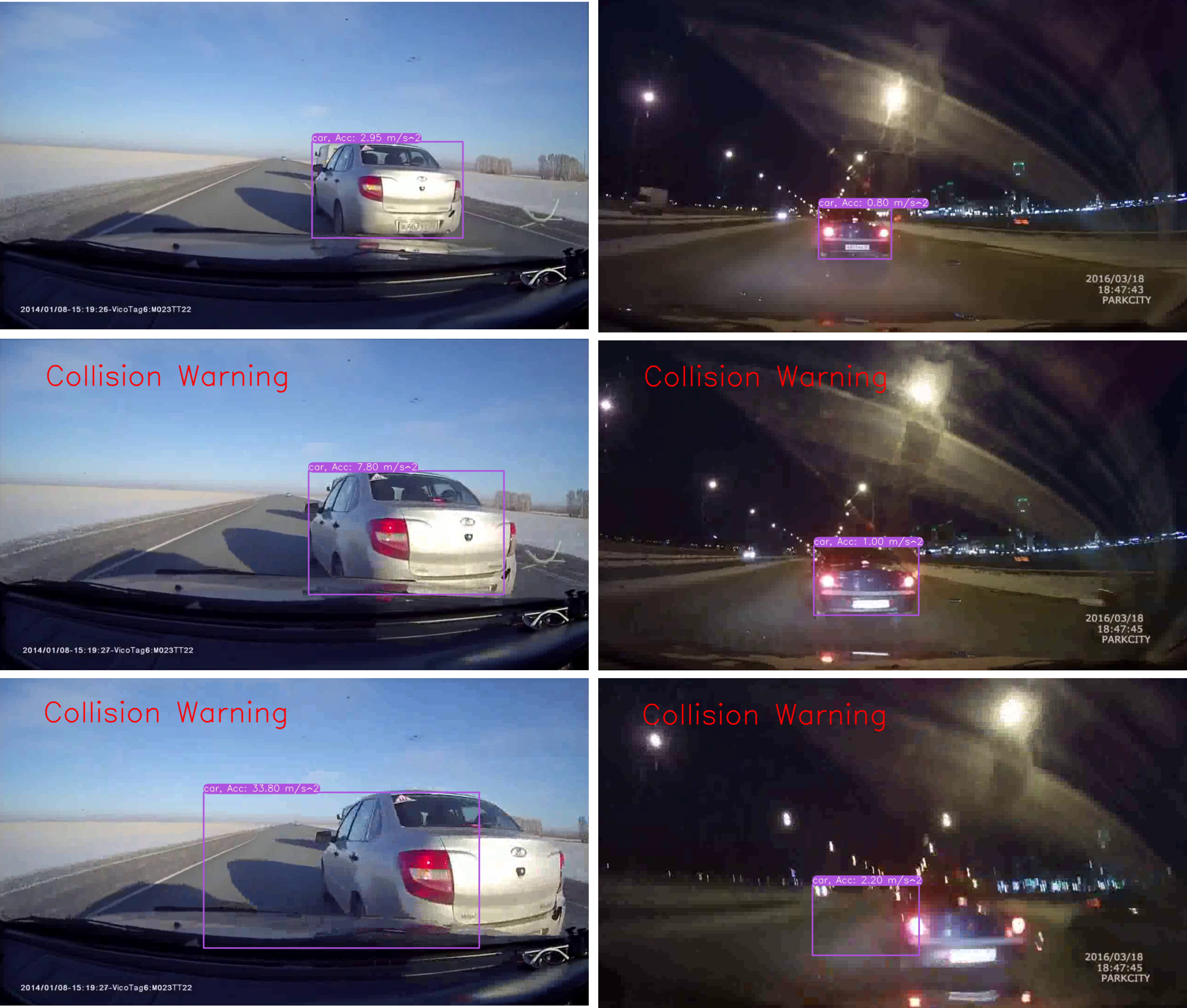}}
\caption{Day and Night Collision Prediction using VCAS on Car Crash Dataset}
\label{fig}
\end{figure}

Fig 2 illustrates real-time parallel OD on three vehicle-mounted dash cam streams having 1920 x 1080 resolution in both day and night. The advantage of having multiple camera streams to get a better understanding of scene is clearly visible. Few objects which are missed with the middle camera (monocular approach) are detected with either the left or right side camera’s view. We have evaluated our own created dataset to see the performance of object detector-tracker results as well as rate of acceleration calculation. For collision prediction analysis, Car Crash Dataset, having actual collision incidents was used, since it was not possible to create actual collision scenario on ground. Fig 3 is a depiction of the collision prediction of V-CAS on the CCD Crash-1500 subset in both day and night. From left to right we can see how the object was detected first, then its relative rate of acceleration and trajectory was continuously calculated, and based upon that a collision warning was generated on screen proactively. On an average of 1.13 seconds in both day and night combined results, proactive alert has been generated before the collision. The cost of complete end-to-end V-CAS solution is around 1200 - 1500 USD making it ideally affordable for auto-manufacturers.

\begin{table*}[htbp]
\caption{V-CAS Overall Performance Evaluation on the Car Crash Dataset}
\begin{center}
\resizebox{\textwidth}{!}{ % Adjust to text width
\begin{tabular}{|c|c|c|ccc|ccc|c|c|}
\hline
\textbf{Category} & \textbf{Total} & \textbf{Ground} & \multicolumn{3}{|c|}{\textbf{V-CAS without Brake Detection}} & \multicolumn{3}{|c|}{\textbf{V-CAS with Brake Detection}} & \textbf{Nvidia RTX} & \textbf{Jetson Orin} \\
\cline{4-9} 
 &  & \textbf{Truth} & \textbf{Predicted} & \textbf{Precision} & \textbf{Accuracy} & \textbf{Predicted} & \textbf{Precision} & \textbf{Accuracy} & \textbf{4090Ti (FPS)} & \textbf{Nano (FPS)} \\
\hline
\textbf{Day} & 1062 & 764 & 759 & 98.68\% & 97.64\% & 760 & 98.94\% & 98.12\% & 62 & 15.6 \\
% \hline
\textbf{Night} & 438 & 376 & 304 & 89.47\% & 68.95\% & 352 & 97.72\% & 90.87\% & 61.8 & 15.1 \\
\hline
\end{tabular}
}
\label{tab:comparison}
\end{center}
\end{table*}

Table V shows the performance of V-CAS from the Crash-1500 subset of CCD. Out of 1500 crash videos, 1140 are the actual crash incidents where the collision occurred to host vehicles themselves. It comprises 764 day and 376 night videos. The day's crash incidents show very promising results with above 98\% precision and almost equivalent accuracy. However, for nighttime videos, due to loss in detection and tracking, there is a drop in accuracy when being used without the brake detection method. However, with incorporating the brake detection module as well, the night performance has remarkably raised to above 90\% accuracy due to a smaller number of FN. The performance on our embedded system is almost real-time with above 15 fps using detections on alternate frames. A slight drop in fps at nighttime is due to a struggle in tracking of objects due to lightning conditions and losing object detection at some points. A final accuracy of 98.12\% for daytime and 90.87\% for night-time videos has been achieved.

\section{CONCLUSION}
In this paper, we have proposed a real-time, multicamera, collision avoidance system V-CAS using custom trained vision-based transformer RT-DETR and DeepSORT. They were being compared for their performance and precision along with the integration technique for multicamera streams for a single object detector-tracker solution. RT-DETR is a balanced choice between inference speed and precision whereas DeepSORT is best for real-time multi-object tracking in diverse scenarios. Our proposed system showed promising results on the Car Crash Dataset in daytime scenarios with above 98\% and 90\% accurate results in daytime and nighttime scenarios. A combination of Brake light detection was used to further enhanced night time performance and robustness of our model. Further quantization of our backbone OD model RT-DETR or latest compression like pruning or knowledge distillation without compromising its accuracy and precision will make the performance even better for embedded systems. Our proposed system is quite precise, computationally efficient, and low-cost real-time solution systems that can be implemented on low-power-embedded platforms for vehicles in everyday life.

\end{document}